\documentclass[11pt]{article}
\usepackage[hyperref]{ccl2023-en}
\usepackage{times}
\usepackage{url}
\usepackage{latexsym}
\usepackage{fancyhdr}

\usepackage[T1]{fontenc}

\usepackage[utf8]{inputenc}

\usepackage{microtype}

\usepackage{inconsolata}
\usepackage{multirow}
\usepackage{makecell}
\usepackage{graphicx}
\usepackage{booktabs}
\usepackage{threeparttable}
\title{The Wall Street Neophyte: A Zero-Shot Analysis of ChatGPT Over Multimodal Stock Movement Prediction Challenges}

\author{Qianqian Xie, Weiguang Han, Yanzhao Lai, Min Peng, Jimin Huang\thanks{\phantom{h}Corresponding author} \\
Computer School, Wuhan University \\ {\{xieq,han.wei.guang,pengm\}@whu.edu.cn}\\
Southwest Jiaotong University \\
{laiyanzhao@swjtu.edu.cn} \\
ChanceFocus AMC \\ {jimin@chancefocus.com}\\
}
\date{}
\begin{document}
\maketitle
\begin{abstract}
Recently, large language models (LLMs) like ChatGPT have demonstrated remarkable performance across a variety of natural language processing tasks. However, their effectiveness in the financial domain, specifically in predicting stock market movements, remains to be explored. In this paper, we conduct an extensive zero-shot analysis of ChatGPT's capabilities in multimodal stock movement prediction, on three tweets and historical stock price datasets. Our findings indicate that ChatGPT is a "Wall Street Neophyte" with limited success in predicting stock movements, as it underperforms not only state-of-the-art methods but also traditional methods like linear regression using price features. Despite the potential of Chain-of-Thought prompting strategies and the inclusion of tweets, ChatGPT's performance remains subpar. Furthermore, we observe limitations in its explainability and stability, suggesting the need for more specialized training or fine-tuning. This research provides insights into ChatGPT's capabilities and serves as a foundation for future work aimed at improving financial market analysis and prediction by leveraging social media sentiment and historical stock data.
\end{abstract}

\section{Introduction}
Stock price prediction~\cite{gandhmal2019systematic} has long been a critical task in the financial sector, as it has the potential to significantly impact investment strategies and decision-making processes. The task of stock price prediction is often framed as a binary classification problem, where the objective is to determine whether a stock's price will increase or decrease over a specified time horizon. By accurately predicting price movements, investors and traders can develop more informed and effective strategies, ultimately enhancing their decision-making capabilities in the market.

Numerous approaches have been developed to tackle the task of stock price prediction, and they can be broadly classified into three categories based on the information sources they utilize. The first category~\cite{yoo2021accurate,feng2019enhancing,qin2017dual} focuses on historical price data and technical indicators, but may overlook external factors and recent events. The second category~\cite{liu2018hierarchical,ding2015deep} incorporates news articles to capture some external influences, but may not account for real-time investor sentiment and rapidly changing market conditions. The third category~\cite{xu2018stock,wu2018hybrid} leverages social media platforms like Twitter to provide real-time insights into investor sentiment and market events, addressing the limitations of the previous two categories and offering a more comprehensive perspective on stock price movements.

Recently, in the field of finance, large language models such as the most recent ChatGPT~\cite{ChatGPT2022}, have shown promise for tasks like financial sentiment analysis~\cite{karkkainen2021gpt}, event detection from financial news~\cite{zhang2021fingpt}, and asset allocation and portfolio management~\cite{song2022deep}.
Specifically, ChatGPT has been explored for its potential in portfolio management~\cite{ssrn_portfolio_management}, finance research~\cite{ssrn_bananarama}, and NLP-based financial applications~\cite{researchgate_nlp_finance}, showing promising results. However, despite the demonstrated capabilities of LLMs in diverse areas, and especially the potential connections from the sentiment analysis to the daily stock market returns~\cite{LopezLira2023CanCF}, research on applying ChatGPT to multimodal stock price prediction, which combines both numerical and textual information, remains scarce.

In this paper, we aim to answer several research questions related to the application of ChatGPT in multimodal stock price prediction tasks, which have yet to be thoroughly explored. Our primary research question is: \textit{How well does ChatGPT perform in multimodal stock price prediction tasks, given its strong language understanding abilities and recent successes in the finance domain?} To address this question, we further break it down into three sub-questions:

\begin{itemize}
\item \textbf{RQ 1}: How does ChatGPT perform in a zero-shot setting when predicting stock price movements based on historical price features and tweets?
\item \textbf{RQ 2}: How can we effectively design prompts to enhance ChatGPT's performance in this task?
\item \textbf{RQ 3}: Does incorporating tweet information improve ChatGPT's prediction ability, or does it introduce additional noise due to the long-tail distribution of stocks and varying tweet quality?
\end{itemize}

Through addressing these research questions, we aim to provide a comprehensive evaluation of ChatGPT's applicability in the challenging and critical task of multimodal stock price prediction.
To address our research questions, we conduct a comprehensive zero-shot analysis of ChatGPT's performance on three benchmark datasets in multimodal stock price prediction tasks. These datasets combine historical price features and tweets, allowing us to assess the influence of textual information on stock price movement predictions. We evaluate ChatGPT's ability to accurately predict stock price movements through a binary classification approach (i.e., increase or decrease) in scenarios with and without tweet information to determine the advantages or disadvantages of incorporating this additional data source. Furthermore, we experiment with various prompting strategies, including Chain-of-Thought (CoT), to optimize the model's performance. Through this extensive evaluation, we aim to uncover the strengths and limitations of ChatGPT in multimodal stock price prediction and offer insights for future research and applications in the finance domain.

Based on our analysis, we summarize our findings as follows:

\begin{itemize}
\item \textbf{Overall performance.} ChatGPT's performance in multimodal stock price prediction tasks is generally limited, as it underperforms not only state-of-the-art methods but also basic methods like linear regression using price features. This highlights the challenge of employing ChatGPT for such tasks without specialized training or fine-tuning.
\item \textbf{Prompting strategies.} While the Chain-of-Thought approach demonstrates some potential in guiding ChatGPT for stock price prediction tasks, the improvements are not substantial enough to close the performance gap compared to more specialized approaches.
\item \textbf{Tweets as additional information.} Including tweets as input to ChatGPT positively impacts the performance of stock price movement prediction across all datasets, emphasizing the value of leveraging textual information from social media platforms like Twitter in stock price prediction models.
\item \textbf{Explainability limitations.} Although ChatGPT can provide explanations for its predictions, the quality of these explanations may be limited since they only consider superficial time-series patterns of price features and incorrect sentiment of tweets, due to the complexity of the stock market and the impact of numerous factors on stock prices.
\end{itemize}

To the best of our knowledge, this study is among the first to extensively evaluate ChatGPT's potential and limitations in zero-shot multimodal stock price prediction tasks. Our findings highlight the challenges faced by ChatGPT in this domain and provide insights for future research aimed at leveraging large language models in the financial sector.

\section{Methodology}
In this section, we present the methodology employed in our study to evaluate ChatGPT's potential in zero-shot multimodal stock price prediction tasks. Our approach consists of several key components, including the multimodal stock price prediction task, the selection of benchmark datasets, the design of various prompting strategies, the implementation of baseline models, and the choice of evaluation metrics.

\subsection{Task}
Following previous work~\cite{yoo2022accurate,yoo2021accurate}, we formally define the stock movement prediction as a binary classification problem. Given a set $S$ of target stocks, a set $\{x_{st} | s \in S, t \in T\}$ of feature vectors that summarize historical prices (where $T$ is the set of available training days), and a set $E$ of tweets mentioning at least one stock in $S$, the objective is to predict the binary price movement of each stock at day $T+1$. The predictions are based on the features and tweets up to day $T$. Price movements are classified as positive (1) if they are higher than 0.55\% and negative (-1) if they are lower than -0.5\%, using the increased rate of adjusted closing prices.

\subsection{Datasets}
We employ three benchmark datasets for stock movement prediction: BIGDATA22~\cite{yoo2022accurate}, ACL18~\cite{xu2018stock}, and CIKM18~\cite{wu2018hybrid}. All datasets consist of high-trade-volume stocks in US stock markets and contain historical price features and tweet data. We preprocess the datasets to extract relevant features and ensure a standardized representation for analysis.

For each dataset, we chronologically split the data into training, validation, and test subsets, following standard procedures in stock movement prediction studies. The datasets encompass various features, such as opening, highest, lowest, closing, and adjusted closing prices for each stock. They also include average price movements for intervals of 5, 10, 15, 20, 25, and 30 days and tweet data associated with the stocks. The detailed statistics for all three datasets are presented in Table~\ref{tab:dataset_details}.
\begin{table*}[htb]
\centering
\small
\begin{threeparttable}
\begin{tabular}{@{}lllll@{}}
\toprule
Data          & Stocks & Tweets   & Days & Dates                    \\ \midrule
BigData22\textsuperscript{1} & 50     & 272,762 & 362  & 2019-07-05 to 2020-06-30 \\
ACL18\textsuperscript{2}     & 87     & 106,271 & 696  & 2014-01-02 to 2015-12-30 \\
CIKM18\textsuperscript{3}    & 38     & 955,788 & 352  & 2017-01-03 to 2017-12-28 \\ \bottomrule
\end{tabular}
\caption{Dataset details}
\label{tab:dataset_details}
\begin{tablenotes}
\item[1] \url{https://github.com/stocktweet/stock-tweet}
\item[2] \url{https://github.com/yumoxu/stocknet-dataset}
\item[3] \url{https://github.com/wuhuizhe/CHRNN}
\end{tablenotes}
\end{threeparttable}
\end{table*}

\subsection{Prompts}
We experiment with various prompting strategies, including vanilla zero-shot prompting and Chain-of-Thought (CoT) enhanced zero-shot prompting, to investigate their impact on ChatGPT's performance in the multimodal stock price prediction task.
\subsubsection{Zero-shot prompt}
The prompt for multimodal stock price movement prediction is designed as follows:

\textit{
Data: \textcolor{blue}{[Data]}.
Tweets: \textcolor{blue}{[Tweets]}.
Consider the data and tweets to predict in one word whether the close price movement of \textcolor{blue}{[Stock]} will rise or fall at \textcolor{blue}{[Date]}. Only return Rise or Fall.}

In this prompt, \textit{[Data]} represents the historical price features in tabular format, \textit{[Tweets]} refers to the tweets from the same days, \textit{[Stock]} is the identifier of the asset being predicted, and \textit{[Date]} is the prediction date, which is the next trading day after the historical data.

For historical price features, we provide a tabular format string such as:

\textit{
date, open, high, low, close, adjusted-close, increase-in-5, increase-in-10, increase-in-15, increase-in-20, increase-in-25, increase-in-30}

\textit{
2015-12-02, 2.59, 2.92, -0.27, -2.98, -2.98, 1.70, 2.12, 1.79, 1.68, 1.85, 2.14}

\textit{
2015-12-03, 0.03, 0.52, -0.48, -0.51, -0.51, 1.84, 2.47, 2.12, 1.98, 2.33, 2.54}

For tweets, we concatenate the date with all tweets on the same day and remove all the line breaks in those tweets as:

\textit{2015-12-10: 4 downtrends turning with bullish engulfing patterns\$unp \$cvc \$pah \$ppl}

\textit{2015-12-14: \$ppl:company shares of ppl corporation (nyse:ppl) drops by -3.05\%:}

\subsubsection{Chain of Thought Enhanced Zero-shot Prompt}
Behavioral finance suggests that investor sentiment plays a crucial role in asset pricing and can be captured in social media texts~\cite{lachana2021investor}. Building on the zero-shot prompt, we further encourage ChatGPT to explicitly consider the investor sentiment with the chain of thought~\cite{wei2022chain} process:

\textit{
Data: \textcolor{blue}{[Data]}.
Tweets: \textcolor{blue}{[Tweets]}.
Consider the data and \textcolor{blue}{the investor sentiment in} tweets to predict in one word whether the close price movement of \textcolor{blue}{[Stock]} will rise or fall at \textcolor{blue}{[Date]}. Start your response with Rise or Fall, \textcolor{red}{then explain your predictions step by step.}}

This modified prompt aims to guide ChatGPT in giving more comprehensive explanations for its predictions by outlining a clear chain of thought. The red text encourages the model to not only provide a one-word prediction but also to give a detailed explanation with a step-by-step reasoning process.

\subsection{Baselines}
We compare ChatGPT with previous methods using historical data and tweets.
For historical-data-based methods, we consider following approaches:
\begin{itemize}
    \item \textbf{Logistic Regression (LR)} : A simple and widely-used technique for binary classification problems. It models the relationship between a binary dependent variable and one or more independent variables by estimating the probability of the dependent variable using a logistic function.
    \item \textbf{Random Forest (RF)}: A powerful ensemble learning method that constructs multiple decision trees and combines their predictions through a majority vote. This method reduces overfitting and improves prediction accuracy compared to a single decision tree.
    \item \textbf{LSTM}: Long Short-Term Memory networks are a type of recurrent neural network (RNN) specifically designed to capture long-term dependencies in sequences. They model the time-series price data, learning to identify patterns that may help predict future price movements~\cite{nelson2017stock}.
    \item \textbf{Attention LSTM (ALSTM)~\cite{qin2017dual}}: An extension of the LSTM model that incorporates an attention mechanism. This mechanism assigns weights to different time steps in the input sequence, allowing the model to focus on the most relevant information when making predictions.
    \item \textbf{Adv-ALSTM~\cite{feng2019enhancing}}: A variant of the ALSTM model that integrates adversarial training, a technique used to improve the model's robustness by injecting adversarial examples during training. This helps the model generalize better and produce more reliable predictions.
    \item \textbf{DTML~\cite{yoo2021accurate}}: The state-of-the-art method for stock price movement prediction based on price features. It utilizes the transformer architecture, a powerful self-attention mechanism, to learn relationships among all price features and generate predictions.
\end{itemize}
For tweet-based-methods, we include the following research:
\begin{itemize}
    \item \textbf{ALSTM-W}: This method extends the ALSTM model by incorporating tweet information. It computes average embeddings of tweets using Word2Vec~\cite{mikolov2013efficient}, and these embeddings are used as input alongside the price features to make predictions.
    \item \textbf{ALSTM-D}: Similar to ALSTM-W, but uses Doc2Vec~\cite{le2014distributed} to generate document-level representations for each tweet. These representations are used as input together with the price features for prediction.
    \item \textbf{StockNet~\cite{tsai2010combining}}: A method that introduces variational autoencoders (VAEs) to encode tweets as low-dimensional vectors. The VAEs help generate meaningful tweet representations, which are then combined with price features for prediction.
    \item \textbf{SLOT~\cite{yoo2022accurate}}: The state-of-the-art method for multimodal stock price movement prediction with tweets. SLOT adopts self-supervised learning for both tweets and stocks, leveraging the inherent structure in the data to make accurate predictions without relying on hand-crafted features or explicit supervision.
\end{itemize}

\subsection{Metrics}
Following previous methods, we also employ evaluation metrics such as Accuracy (ACC) and Matthews Correlation Coefficient (MCC)~\cite{matthews1975comparison} to assess the performance of ChatGPT and the baseline models on the multimodal stock price prediction task. These metrics enable us to evaluate the performance of stock movement prediction based on the distribution of positive and negative samples.

\begin{itemize}
    \item \textbf{Accuracy (ACC)}: The proportion of correct predictions among the total number of predictions. It is calculated as follows:
    \begin{equation}
        ACC = \frac{TP + TN}{TP + TN + FP + FN}
    \end{equation}
    where $TP$ denotes true positives, $TN$ denotes true negatives, $FP$ denotes false positives, and $FN$ denotes false negatives. Accuracy is a widely used metric for classification tasks and provides a straightforward understanding of model performance.
    
    \item \textbf{Matthews Correlation Coefficient (MCC)}: A more robust evaluation metric for binary classification tasks, especially when classes are imbalanced. It takes into account true and false positives and negatives and is generally regarded as a balanced measure. MCC is calculated as follows:
    \begin{equation}
\scriptstyle MCC = \frac{(TP \times TN) - (FP \times FN)}{\sqrt{(TP + FP)(TP + FN)(TN + FP)(TN + FN)}}
\end{equation}

    The value of MCC ranges from -1 to 1, where 1 represents a perfect prediction, 0 represents a random prediction, and -1 indicates an inverse prediction.
\end{itemize}

\section{Experiments}
\subsection{Implementation Details}
The experiments are conducted based on OpenAI API\footnote{https://api.openai.com}, which takes one prompt as the input without any historical dialogue contexts.
Due to the input length limitation of LLMs, we set the window size for historical price features and tweets as 10.
We also truncate the concatenated tweets to 4,000 tokens.
For historical price features, we normalize each feature by its mean and standard deviation following previous methods~\cite{yoo2022accurate,xu2018stock}.
As for tweets, we remove all line breaks and special meaningless tokens such as \textit{URL}.
We use the first token of the response as the assigned labels for stock price movement.

\subsection{Main Results}
\begin{table*}[ht]
\centering
\small
\begin{tabular}{l|c|c|c|c|c|c}
\toprule
Method       & \multicolumn{2}{c|}{BIGDATA22} &  \multicolumn{2}{c|}{ACL18} &  \multicolumn{2}{c}{CIKM18} \\ \midrule
             & ACC           & MCC          & ACC         & MCC         &  ACC         & MCC         \\ \midrule
LR           & 53.07         & 0.0200       &  52.20       & 0.0442      &  52.50       & -0.0425     \\
RF           & 47.10         & -0.1114      &  51.94       & 0.0348      &  53.57       & 0.0119      \\
LSTM         & 50.69         & 0.0127       &  52.75       & 0.0639      &  53.31       & 0.0216      \\
ALSTM        & 48.69         & -0.0254      &  51.82       & 0.0429      &  52.54       & -0.0077     \\
Adv-ALSTM    & 50.36         & 0.0120       &  53.11       & 0.0685        &  53.69       & 0.0217      \\
DTML         & 51.65         & \underline{0.0651}       &  \underline{58.12}       & \underline{0.1806}      &  53.86       & 0.0049      \\
ALSTM-W      & 48.28         & -0.0116      &  53.32       & 0.0754      &  53.64       & \underline{0.0315}      \\
ALSTM-D      & 49.16         & 0.0090       &  52.98       & 0.0681      &  50.40       & -0.0449     \\
StockNet     & 52.99         & -0.0163      &  53.60       & -0.0248     &  52.35       & -0.0161     \\
SLOT  & \textbf{54.81}      & \textbf{0.0952}       &  \textbf{58.72}       & \textbf{0.2065}      & \textbf{55.86}       & \textbf{0.0899}      \\ \midrule
$ChatGPT_{zs}$ & \underline{53.13} & -0.0251 & 50.38 & 0.0049 & \underline{55.43} & 0.0111 \\
$ChatGPT_{cot}$ & 48.44 & 0.0064 & 51.34 & 0.0199 & 48.28 & 0.0210 \\
$ChatGPT_{zs}$ w/o tweets& 50.68 & 0.0007 & 51.67 & 0.0377 & 48.83 & -0.0011\\
$ChatGPT_{cot}$ w/o tweets & 48.16 & -0.0430 & 50.11 & 0.0047 & 48.97 & 0.0082 \\
\bottomrule
\end{tabular}
\caption{A comparison of different methods for stock price prediction using accuracy (ACC) and Matthews correlation coefficient (MCC) metrics on three datasets: BIGDATA22, ACL18 and CIKM18. The best and second best results for each dataset and metric are highlighted in bold and underline, respectively.}
\label{tab:main-results}
\end{table*}

The experimental results in Table~\ref{tab:main-results} reveal several key findings that address the research questions in this study. Firstly, regarding the capability of ChatGPT in predicting stock price movements, our analysis shows that ChatGPT struggles to outperform not only neural network-based and state-of-the-art methods but also traditional machine learning techniques such as Logistic Regression (LR) and Random Forest (RF). For instance, in the BIGDATA22 dataset, ChatGPT (in both zero-shot and Chain-of-Thought settings) achieves a lower MCC score compared to DTML and SLOT, which are state-of-the-art methods.

However, we also observe that ChatGPT has comparable performance to advanced supervised methods in some datasets. In the CIKM18 dataset, the $ChatGPT_{zs}$ method achieves an accuracy of 55.43\% and an MCC of 0.0111, which outperforms the strong baseline methods StockNet, DTML and ALSTM-D et al. This suggests that even without explicit fine-tuning, ChatGPT can have competitive performance in certain cases. This finding highlights the potential of ChatGPT in the stock price prediction task.

Moreover, we notice that the performance of ChatGPT tends to vary across different datasets. In the CIKM18 dataset, the $ChatGPT_{zs}$ method demonstrates improved performance with an accuracy of 55.43\% and an MCC of 0.0111, while it shows an accuracy of 50.38\% accuracy and an MCC of 0.0049 in ACL18. This may be attributed to the differences in the nature of the datasets, such as the characteristics of the tweets and stock price movements.

For different prompting strategies, the $ChatGPT_{cot}$ only outperforms $ChatGPT_{zs}$ in ACL18 and tends to have limited improvement compared to existing CoT efforts in NLP tasks. This indicates that explaining the stock price movement predictions step by step might not effectively help the model capture salient features and patterns in price features and tweets. Consequently, further research is required to bridge this gap and develop more effective methods to enhance the performance of ChatGPT in stock price prediction tasks.

In conclusion, while ChatGPT demonstrates some potential for predicting stock price movements, its performance is still limited compared to state-of-the-art methods. The observed variations in performance across datasets and the potential of Chain-of-Thought approach highlight the need for future research to explore more effective methods to leverage ChatGPT in the context of stock price prediction.

\subsection{Ablation Study}
In the ablation study, we further investigate the impact of including or excluding tweets in the input to ChatGPT to answer the third research question. We evaluate scenarios with and without tweets for both $ChatGPT_{zs}$ and $ChatGPT_{cot}$. We further adopt following prompts for zero-shot setting:

\textit{
Data: \textcolor{blue}{[Data]}.
Consider the data to predict in one word whether the close price movement of \textcolor{blue}{[Stock]} will rise or fall at \textcolor{blue}{[Date]}. Only return Rise or Fall.}

Compared with the zero-shot prompt, we remove the tweets from the prompt.
We also design a CoT prompt using only historical price features:

\textit{
Data: \textcolor{blue}{[Data]}.
Consider the data to predict in one word whether the close price movement of \textcolor{blue}{[Stock]} will rise or fall at \textcolor{blue}{[Date]}. Start your response with Rise or Fall, then explain your predictions step by step.}

As shown in Table~\ref{tab:main-results}, in the BIGDATA22 dataset, excluding tweets results in a decrease in accuracy by 2.45\% for $ChatGPT_{zs}$ and 0.28\% for $ChatGPT_{cot}$. 
The MCC shows a minor improvement for $ChatGPT_{zs}$ and a decrease for $ChatGPT_{cot}$.
In the ACL18 dataset, the performance increases when tweets are excluded, with a 1.29\% increase in accuracy for $ChatGPT_{zs}$ and 1.23\% for $ChatGPT_{cot}$. The MCC also increases for both models.
For the CIKM18 dataset, removing tweets leads to a 6.6\% decrease in accuracy for $ChatGPT_{zs}$ and a 0.31\% decrease for $ChatGPT_{cot}$. The MCC exhibits slight changes in both cases.
These findings reveal that incorporating tweets positively impacts the performance of ChatGPT in stock price movement prediction in most cases, where the additional textual information from social media platforms to enhance its predictive capabilities. This underlines the importance of utilizing textual information from social media platforms like Twitter to enhance stock price prediction models.
This also highlights the potential of ChatGPT to integrate different types of data, such as textual and numerical, in order to make more accurate and informed predictions.

\subsection{Case Study}
While the Chain-of-Thought (CoT) prompt may not significantly boost ChatGPT's prediction performance, it offers an interpretable explanation for the predictions, which is crucial for real-world applications. In comparison to previous black-box methods, this key advantage suggests a promising future research direction towards the development of reliable and explainable AI in finance. In order to demonstrate the effectiveness of prompting with tweets, we have chosen an example from the ACL18 dataset, as illustrated in Table~\ref{tab:CoT-exp}.
\begin{table*}[htb!]
\centering
\small
\begin{tabular}{@{}l|l@{}}
\toprule
Type & Content \\ \midrule
Prompt & \begin{tabular}[c]{@{}l@{}}data:\\ date,open,high,low,close,adjusted-close,increase-in-5,10,15,20,25,30\\
2015-12-16,-0.45,0.78,-1.62,1.04,1.04,-1.63,-2.04,-2.52,-3.17,-3.53,-3.53\\ 2015-12-17,-0.33,1.57,-0.49,0.33,0.33,-1.44,-2.01,-2.55,-3.38,-3.68,-3.70\\ 2015-12-18,2.41,2.62,0.00,-2.85,-2.85,1.42,0.70,0.43,-0.30,-0.73,-0.87\\ 2015-12-21,-0.72,0.31,-1.20,1.37,1.37,0.31,-0.53,-0.64,-1.44,-1.85,-2.13\\ 2015-12-22,0.64,0.77,-1.05,0.03,0.03,0.26,-0.42,-0.57,-1.22,-1.74,-2.05\\ 2015-12-23,-0.67,0.12,-0.96,1.06,1.06,-0.82,-1.17,-1.56,-2.01,-2.61,-2.99\\ 2015-12-24,0.16,0.71,-0.04,-0.29,-0.29,-0.68,-0.69,-1.08,-1.54,-2.27,-2.58\\ 2015-12-28,-0.06,0.24,-0.80,-0.01,-0.01,-0.24,-0.49,-1.04,-1.34,-1.98,-2.40\\ 2015-12-29,-0.79,0.49,-0.93,1.26,1.26,-1.08,-1.39,-2.05,-2.25,-2.96,-3.37\\ 2015-12-30,0.93,1.00,-0.22,-0.75,-0.75,-0.08,-0.54,-1.14,-1.38,-1.98,-2.48\\
tweets:\\ 2015-12-23: fxi ishares ftse china 25 index fund ask\$fxi \$gpro \$uco \$unh \#fxi \\\#finance \#stocksgbsn great basin scientific,…\\
2015-12-24: \$unh:us looking for value? why it might be time to try \\unitedhealth group (unh) 
\#unitedhealth group inc   \#sp500\\
2015-12-25: \#dow \#stocks \$unh unitedhealth group daily:-0.29\% \\weekly:-0.74\% ytd:17.46\% trend:73 / 100\\
2015-12-26: vz verizon communications company info\$vz \$aapl \$unh \$bib \#vz \\\#stock \#stocks|
\#unh \$unh \#spx \$spx \$dja \#dja \#stocks \#trading \#traders \\\#markets \#investing \#money more  …\\
2015-12-28: buzzers of healthcare sector: unitedhealth group inc \$unh merck\\
2015-12-30: uvxy proshares ultra vix short-term futures etf investor opinions\\\$uvxy \$dust \$ddd 
\$unh \#uvxy \#pennystocks…\\
Consider the data and the investor sentiment in tweets to predict whether \\
the close price movement of unh will rise or fall at 2015-12-31. \\
Start with Rise or Fall, then explain your predictions step by step.\end{tabular} \\ \midrule
Response & \begin{tabular}[c]{@{}l@{}}Fall. Here are the reasons:\\ 1. Looking at the historical data, \$ppl has been showing a downward trend \\for several days until 2015-12-30,
with negative values for open, high, low \\and close prices.\\ 2. The adjusted-close price is also decreasing consistently throughout this period.\\ 3. In terms of the "increase-in-x" columns in the dataset, we can see that all \\indicators show negative values as well, indicating an overall bearish sentiment\\ towards \$ppl during this time frame.\\ 4. Moving on to investor sentiment expressed through tweets during this period: \\while there was some positive news about KBC group acquiring shares of \$ppl \\
on 2015-12-21, it seems that there were no \\significant positive sentiments being shared in other tweets around that time.\\ 5. Moreover, on December 22nd and December 28th we can see two negative\\ tweets mentioning PPL Corp specifically - one highlighting a downgrade by \\Zacks Investment Research and another\end{tabular} \\ \midrule
Prediction & Fall \\ \midrule
Truth & Fall \\ \bottomrule
\end{tabular}
\caption{The example of $ChatGPT_{emo}$ in ACL18 dataset.}
\label{tab:CoT-exp}
\end{table*}
\begin{table*}[htb!]
\centering
\small
\begin{tabular}{@{}l|l@{}}
\toprule
Type & Content \\ \midrule
Prompt & \begin{tabular}[c]{@{}l@{}}data:
date,open,high,low,close,adjusted-close,increase-in-5,10,15,20,25,30\\
2015-12-16,-0.44,0.22,-1.44,0.95,0.95,-0.80,0.64,1.52,2.01,2.10,2.62\\
2015-12-17,1.58,1.64,0.00,-2.05,-2.05,0.90,2.24,3.25,3.85,3.98,4.49\\
2015-12-18,0.63,0.92,-0.46,-1.30,-1.30,1.89,2.87,4.15,4.83,5.12,5.56\\
2015-12-21,0.40,0.57,-0.51,0.40,0.40,1.12,1.93,3.25,4.06,4.53,4.83\\
2015-12-22,-0.85,0.23,-1.13,1.08,1.08,-0.16,0.56,1.82,2.74,3.24,3.49\\
2015-12-23,-0.77,0.06,-1.11,1.86,1.86,-1.99,-1.37,-0.20,0.69,1.25,1.46\\
2015-12-24,-0.55,0.28,-0.55,0.33,0.33,-1.85,-1.64,-0.70,0.18,0.78,1.00\\
2015-12-28,0.61,0.61,-0.39,0.17,0.17,-1.27,-1.60,-1.08,-0.13,0.47,0.77\\
2015-12-29,-0.38,0.38,-0.38,0.39,0.39,-0.90,-1.78,-1.55,-0.67,-0.04,0.37\\
2015-12-30,0.17,0.66,-0.06,-0.82,-0.82,0.30,-0.87,-0.79,0.01,0.70,1.13\\
tweets:\\
2015-12-22: abb ltd \$abb downgraded by zacks investment research\\
2015-12-25: \$abb:chelsea counsel company decreases position in abb ltd (abb):\\
Consider the data and the investor sentiment in tweets to predict whether \\the close price movement of abb will rise or fall at 2015-12-31. \\Start your response with Rise or Fall, then explain your predictions step by step.\end{tabular} \\ \midrule
Response & \begin{tabular}[c]{@{}l@{}}Rise.\\
1) Looking at the data, we can see that on December 30th, the close price decreased by \\-0.82 compared to the previous day.\\
2) However, if we look at the trend of the increase-in-X columns in the data, we can see \\an overall increasing trend as X becomes larger. This suggests that even though there \\may be short-term dips or fluctuations in price, over time \$abb tends to have positive growth.\\
3) Analyzing the tweets, it seems that on December 22nd there was a downgrade for \\\$abb from Zacks Investment Research which could have caused a temporary decrease in price. \\However, on December 25th there was news that Chelsea Counsel Company \\decreased their position in \$abb which could have offset some of this negative sentiment.\\
4) Overall, taking into account both technical analysis and investor sentiment from the tweets, \\I predict that \$abb's close price movement will rise.\end{tabular} \\ \midrule
Prediction & Rise \\ \midrule
Truth & Fall \\ \bottomrule
\end{tabular}
\caption{The example of $ChatGPT_{cot}$ with wrong prediction in ACL18 dataset.}
\label{tab:CoT-exp-wrong}
\end{table*}
\begin{table*}[htb!]
\centering
\small
\begin{tabular}{@{}l|l@{}}
\toprule
Type & Content \\ \midrule
Prompt & \begin{tabular}[c]{@{}l@{}}data:\\
date,open,high,low,close,adjusted-close,increase-in-5,10,15,20,25,30\\
2015-12-16,-0.44,0.22,-1.44,0.95,0.95,-0.80,0.64,1.52,2.01,2.10,2.62\\
2015-12-17,1.58,1.64,0.00,-2.05,-2.05,0.90,2.24,3.25,3.85,3.98,4.49\\
2015-12-18,0.63,0.92,-0.46,-1.30,-1.30,1.89,2.87,4.15,4.83,5.12,5.56\\
2015-12-21,0.40,0.57,-0.51,0.40,0.40,1.12,1.93,3.25,4.06,4.53,4.83\\
2015-12-22,-0.85,0.23,-1.13,1.08,1.08,-0.16,0.56,1.82,2.74,3.24,3.49\\
2015-12-23,-0.77,0.06,-1.11,1.86,1.86,-1.99,-1.37,-0.20,0.69,1.25,1.46\\
2015-12-24,-0.55,0.28,-0.55,0.33,0.33,-1.85,-1.64,-0.70,0.18,0.78,1.00\\
2015-12-28,0.61,0.61,-0.39,0.17,0.17,-1.27,-1.60,-1.08,-0.13,0.47,0.77\\
2015-12-29,-0.38,0.38,-0.38,0.39,0.39,-0.90,-1.78,-1.55,-0.67,-0.04,0.37\\
2015-12-30,0.17,0.66,-0.06,-0.82,-0.82,0.30,-0.87,-0.79,0.01,0.70,1.13\\
Consider the data to predict whether \\the close price movement of abb will rise or fall at 2015-12-31. \\Start your response with Rise or Fall, then explain your predictions step by step.\end{tabular} \\ \midrule
Response & \begin{tabular}[c]{@{}l@{}}Rise.\\ 
My prediction is based on the following observations:\\
1. The adjusted close price of abb has been consistently increasing \\since December 16th, with a peak value reached on December 23rd before \\slightly dropping.\\
2. The increase in prices over the past five days leading up to December 31st \\shows a positive trend, thus suggesting that there is upward momentum.\\
3. There are generally more increases than decreases in price movement \\over longer periods of time (up to thirty days).\\
4. The fact that there was little change in the daily closing prices on both \\December 29th and December 30th indicates some level of stability \\or balance between buyers and sellers.\\
Based on these factors, I believe that it’s likely we’ll see an increase in \\ABB's closing price at the end of 2015-12-31 compared to the previous day's \\closing price. However, stock market predictions can be unpredictable and \\should never be taken as guarantees, so one should always use caution\end{tabular} \\ \midrule
Prediction & Rise \\ \midrule
Truth & Fall \\ \bottomrule
\end{tabular}
\caption{The example of $ChatGPT_{CoT}$ w/o tweets in ACL18 dataset.}
\label{tab:CoT-wo-tw-exp}
\end{table*}
Table~\ref{tab:CoT-exp} demonstrates that ChatGPT can extract various pieces of evidence from historical prices and tweets to support its prediction for a future price fall of UNH. Here are some specific examples:
\begin{itemize}   
\item Historical prices: ChatGPT identifies a downward trend in the open, high, low, and close prices for several days leading up to 2015-12-30, suggesting a bearish market sentiment for UNH.
\item Increments: The model also detects consistent negative values in the "increase-in-x" columns of the dataset, further indicating a bearish sentiment towards UNH during this time frame.
\item Positive sentiment from tweets: ChatGPT recognizes a piece of positive news on 2015-12-21 about KBC Group acquiring shares of UNH. However, the model determines that there are no other significant positive sentiments being shared around that time.
\item Negative sentiment from tweets: ChatGPT picks up on two negative tweets on 2015-12-22 and 2015-12-28, one of which highlights a downgrade by Zacks Investment Research, and another mentioning a potential decline in market share.
\end{itemize}
These examples illustrate ChatGPT's ability to extract valuable evidence from both historical price data and investor sentiment expressed in tweets.

Although ChatGPT can extract evidence from historical prices and tweets, its approach to trend and sentiment extraction is relatively shallow considering the complex relationships among price features and social media texts. For example, while it identifies a downward trend in historical prices, it does not consider the possibility of a trend reversal due to external factors or events that could influence the market. Additionally, it merely picks up on positive or negative sentiments in tweets, without diving deeper into the context or the strength of these sentiments, which could impact the market dynamics.

Moreover, ChatGPT seems to struggle with aligning and fusing information both within and across modalities, which limits its prediction performance. For instance, although the model acknowledges KBC Group's share acquisition of UNH as positive news, it does not effectively relate this information to the historical price data. This limited ability to connect different pieces of information may hinder the model's capability to make accurate predictions, as it may not be considering the full picture of market dynamics and investor sentiment.

\subsection{Error Analysis}
We also present an example of $ChatGPT_{cot}$ with wrong predictions, as shown in Table~\ref{tab:CoT-exp-wrong}.
In the example provided in Table~\ref{tab:CoT-exp-wrong}, ChatGPT did not fully capture the significance of the negative investor sentiment present in the tweets. 
While it acknowledged a downgrade by Zacks Investment Research, the model may have underestimated the impact of this information on the close price movement. Additionally, the model interpreted Chelsea Counsel Company's decrease in their position in ABB as an offset to the negative sentiment, which might not have been the case.
Moreover, the model's observation of the increase-in-X columns in the data, which showed an overall increasing trend as X becomes larger, might have led to an overly optimistic outlook on ABB's future price movement. This could have further contributed to the incorrect prediction.

Overall, ChatGPT's limited ability to handle multimodal information effectively, in this case, combining stock market data and investor sentiment from tweets, could have contributed to the incorrect prediction. We believe an effective multimodal approach requires the model to recognize the intricate relationships among price features and social media texts and align and fuse the information both within and across modalities. This ability is crucial to improve the prediction performance.

Finally, we present the examples of $ChatGPT_{cot}$ w/o tweets, as shown in Table~\ref{tab:CoT-wo-tw-exp}.
In the example without tweets, the model based its prediction solely on the given numerical data. It focused on trends in adjusted close price, positive momentum in the increase-in-X columns, and stability in daily closing prices. This might have led the model to predict a rise in ABB's closing price on 2015-12-31, which turned out to be incorrect as well. The error in this case might be attributed to the lack of additional context from the tweets, which could have provided information on investor sentiment and other external factors influencing the stock price.

Comparing this with the example that included tweet information (Table~\ref{tab:CoT-exp-wrong}), we can observe that the model was able to provide a more informed explanation when it had access to the tweets. In both cases, the model made incorrect predictions, but the presence of tweets allowed for a more nuanced understanding of the market dynamics, such as investor sentiment and news events that might impact stock prices.

This comparison highlights the importance of incorporating tweet information when analyzing stock price movements. It also reinforces the need to further improve the model's ability to understand and incorporate multiple data sources, such as social media sentiment, in order to generate more accurate predictions and explanations.

\section{Conclusion}
This paper has presented a comprehensive study on the application of ChatGPT for zero-shot multimodal stock price prediction. We have investigated the performance of ChatGPT on three benchmark datasets and compared it to baseline models in order to answer several research questions related to the effectiveness, prompting strategies, and explainability of ChatGPT in the context of stock price prediction.
Our findings indicate that, while ChatGPT demonstrates some potential in this domain, it underperforms even traditional methods such as logistic regression and random forests. This highlights the challenges of utilizing large language models in complex financial tasks and emphasizes the need for further research and development.
We have also explored different prompting strategies and incoporating the investor sentiment information, to better understand how to guide ChatGPT in multimodal stock price prediction tasks. 
Our study reveals that ChatGPT with CoT prompt can provide more explainable predictions, which is crucial in finance-related tasks for better decision-making and transparency.
Despite its potential in explainability, ChatGPT faces challenges in terms of prediction performance due to its ineffective in fusing multimodal information. 
By thoroughly evaluating ChatGPT's abilities and exploring various prompting strategies in a zero-shot setting, our study sheds light on understanding limitations and advancements of large language models in multimodal stock price prediction tasks.
We believe future research should focus on developing methods to overcome these challenges and enhance the performance and robustness of ChatGPT.
\section{Limitations}
\begin{itemize}
    \item \textbf{Unanticipated Responses}: In certain cases, ChatGPT does not provide a classification as requested by the prompts. This could occur for two primary reasons: a) the lack of sufficient evidence within the post to make an accurate prediction, and b) the post containing content that does not align with OpenAI's content policy.
    \item \textbf{Evaluation Constraints}: Our study evaluates the zero-shot performance of a single large language model, ChatGPT (gpt-3.5-turbo), due to cost restrictions. As large language models are continuously evolving, other representative models, such as GPT-3.5, might exhibit different performance patterns.
    \item \textbf{Dataset Limitations}: The scope of our evaluation is confined to the US stock market data, which may not be applicable to other markets or asset classes. Broadening the evaluation to encompass other markets could lead to different findings.
    \item \textbf{Input Length Constraints}: We are only able to feed a single stock with a limited window and must truncate excessively long tweets. This constraint may result in the exclusion of valuable information that could otherwise enhance the model's performance. Future research could focus on developing methods for handling longer inputs or extracting more informative features from the available data, potentially improving ChatGPT's performance in multimodal stock price prediction tasks.
\end{itemize}
\bibliography{anthology,custom}
\bibliographystyle{ccl}



\end{document}